\documentclass[letterpaper]{article} %
\usepackage{aaai24}  %
\usepackage{times}  %
\usepackage{helvet}  %
\usepackage{courier}  %
\usepackage[hyphens]{url}  %
\usepackage{graphicx} %
\usepackage{natbib}  %
\usepackage{caption} %
\usepackage{algorithm}
\usepackage{algorithmic}

\usepackage{newfloat}
\usepackage{listings}
\DeclareCaptionStyle{ruled}{labelfont=normalfont,labelsep=colon,strut=off} %
\floatstyle{ruled}
\newfloat{listing}{tb}{lst}{}
\floatname{listing}{Listing}
\usepackage{tabularx}
\usepackage{todonotes}
\usepackage{multirow}
\usepackage{graphicx}
\usepackage{enumitem}
\usepackage{anyfontsize}
\usepackage{url}
\usepackage{subcaption}
\usepackage{xr}
\usepackage{bbold}
\usepackage{booktabs}
\usepackage{xcolor,pifont}
\usepackage{MnSymbol,wasysym}
\usepackage{fontawesome}

\usepackage{times}
\usepackage{latexsym}

\usepackage{xspace}
\usepackage{caption}
\usepackage{array}
\usepackage{tcolorbox}
\usepackage{xcolor,colortbl}
\usepackage{makecell}
\usepackage{soul}
\usepackage{stackengine}
\newcolumntype{H}{>{\setbox0=\hbox\bgroup}c<{\egroup}@{}}
\newcommand{\vtwo}[1]{#1}

\newcommand{\Skip}[1]{}

\newcommand{\modelname}{\textsc{Star}\xspace}

\newcommand{\ie}{\textit{i}.\textit{e}.\ }
\newcommand{\eg}{\textit{e}.\textit{g}.\ }

\newcommand{\secref}[1]{\S\ref{#1}}

\newcommand{\figref}[1]{Figure~\ref{#1}}
\newcommand{\tbref}[1]{Table~\ref{#1}}

\newcommand{\dotieconcat}[2]{%
  \text{\raisebox{.8ex}{$\smallfrown$}}%
}
\newcommand{\mypar}[1]{\paragraph{#1}}

\title{STAR: Boosting Low-Resource Information Extraction by Structure-to-Text Data Generation with Large Language Models}
\author{
    Mingyu Derek Ma$^{1}$,
    Xiaoxuan Wang$^{1}$,
    Po-Nien Kung$^{1}$\\
    {\bf P. Jeffrey Brantingham}$^{2}$,
    {\bf Nanyun Peng}$^{1}$,
    {\bf Wei Wang}$^{1}$
}
\affiliations{

    $^{1}$Department of Computer Science, University of California, Los Angeles
    \\
    $^{2}$Department of Anthropology, University of California, Los Angeles
    \\
    \{ma, xw27, ponienkung, violetpeng, weiwang\}@cs.ucla.edu,
    branting@ucla.edu
}

\begin{document}

\maketitle

\begin{abstract}
    \vtwo{Information extraction} tasks such as event extraction require an in-depth understanding of the output structure and sub-task dependencies. They heavily rely on task-specific training data in the form of (passage, target structure) pairs to obtain reasonable performance. However, obtaining such data through human annotation is costly, leading to a pressing need for low-resource information extraction approaches that require minimal human labeling for real-world applications.
\vtwo{
Fine-tuning supervised models with synthesized training data would be a generalizable method, but the existing data generation methods either still rely on large-scale ground-truth data or cannot be applied to complicated IE tasks due to their poor performance.
}
To address these challenges, we propose \modelname, a data generation method that leverages Large Language Models (LLMs) to synthesize data instances given limited seed demonstrations, thereby boosting low-resource information extraction performance.
Our approach involves generating target structures ($Y$) followed by generating passages ($X$), all accomplished with the aid of LLMs. We design fine-grained step-by-step instructions to obtain the initial data instances. 
We further reduce errors and improve data quality through self-reflection error identification and self-refinement with iterative revision.
Our experiments show that the data generated by \modelname significantly improve the performance of low-resource event extraction and relation extraction tasks, even surpassing the effectiveness of human-curated data. 
\vtwo{Human assessment of the data quality shows \modelname-generated data exhibit higher passage quality and better align with the task definitions compared with the human-curated data.}

\end{abstract}

\section{Introduction}
\label{sec:intro}

\vtwo{Information extraction (IE) aims to extract knowledge of certain perspectives from natural language and consolidate it into an output structure~\cite{Ma2021EventPlusTemporalEvent}.
To induce the target structure, the IE models need to understand fine-grained task requirements and constraints. Taking event extraction (EE), which is a component for IE systems to identify event triggers, event types, and their related details as arguments, as an example}, task-specific rules include the predicted spans should be subsequences of the input passage, and arguments should be participants or attributes of the event. EE models are also expected to be aware of the dynamic skeleton of the event structure because the different predicted event types result in their respective sets of argument roles being filled in.
Supervised models learn the implicit requirement and ontology knowledge from training data in the form of (passage, target structure) pairs~\cite{Ma2021HyperExpanTaxonomyExpansion}. Prompt-based inference-only approaches with Large Language Models (LLMs) are shown to be unable to \vtwo{solve these complicated IE tasks}~\cite{Li2023EvaluatingChatGPTInformation,Gao2023ExploringFeasibilityChatGPT,Han2023InformationExtractionSolved}.
In real-world applications, text from various sources and domains contains a broad range of \vtwo{output spaces and label definitions}.
\vtwo{It is costly and rigid to annotate sufficient training data, thus, performing IE given minimal seed data instances is of particular interest for realistic IE applications.}\looseness=-1

The limited resources make understanding the task requirement and structure skeleton even harder.
Existing low-resource IE methods mostly leverage other tasks via transfer learning~\cite{Huang2018ZeroShotTransferLearning,Zhang2021ZeroshotLabelAwareEvent} or reformulate the task into alternative data-rich tasks for indirect supervision~\cite{Sainz2022TextualEntailmentEvent,Xu2023CanNLIProvidea,Lu2022SummarizationIndirectSupervision}.
These works heavily depend on the availability of the source tasks' data and the compatibility between tasks, limiting its application to richer and broader \vtwo{output label spaces} and under-represented domains. 

\vtwo{
Synthesizing additional training data to fine-tune supervised models would be a generalizable method and has demonstrated its success for tasks like sentiment analysis~\cite{Ye2022ZeroGenEfficientZeroshot} and relation extraction \cite{Josifoski2023ExploitingAsymmetrySynthetic}. Some works annotate unlabeled examples with existing models as weak annotators ($X_{gold} \rightarrow Y$)~\cite{He2021GenerateAnnotateLearn,Chia2022RelationPromptLeveragingPromptsa}, produce analogous input ($X_{gold} \rightarrow X^\prime$)~\cite{Kumar2020DataAugmentationUsing,lee2021neural}, or generate input assuming the labels are available ($Y_{gold} \rightarrow X$) \cite{Meng2022GeneratingTrainingData,Gao2021MakingPretrainedLanguage,Josifoski2023ExploitingAsymmetrySynthetic}.
However, these works require ground-truth $X_{gold}$ or $Y_{gold}$, limiting their generalizability and scalability.
What's more, they are designed for classification or straightforward IE tasks, and the performance drops significantly as the task complexity increases. 
Applying them to complicated IE tasks with sub-task dependencies and dynamic output structures such as EE produces noisy data that may impair task performance. 
}

\begin{figure*}[h]
    \centering
    \includegraphics[width=\textwidth]{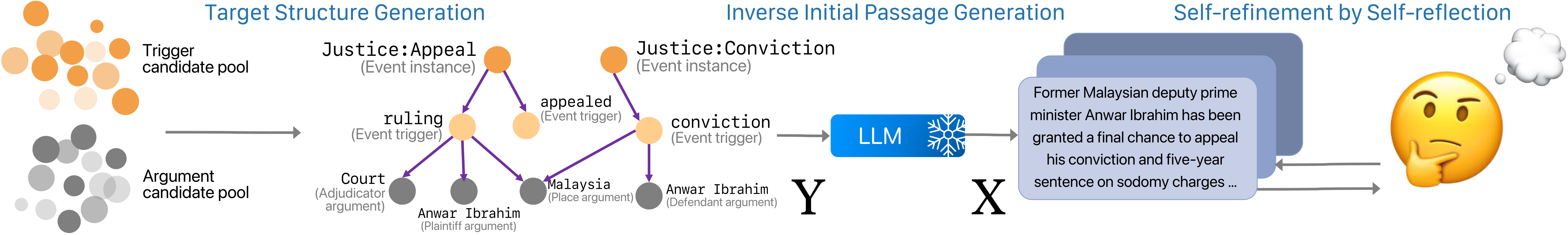}
    \caption{The \modelname inverse data generation strategy using event extraction task as an example. We first generate target structures from valid trigger and argument candidates. Then we prompt the LLM with task instructions from different task granularities to generate the initial passage $X_0$ containing the event information in the given target structure $Y$. Finally, we create self-reflection questions to prompt LLM to identify quality issues automatically and refine the passage with template-based hindsight feedback.
    }
    \label{fig:reverse_gen}
\end{figure*}

In this paper, we present \modelname, a \textbf{S}tructure-to-\textbf{T}ext Dat\textbf{A} Gene\textbf{R}ation pipeline to produce dependable data instances for low-resource IE. 
Instead of using existing models to produce silver target structures $Y$ derived from the input passages $X$ (\ie $X_{gold} \rightarrow Y$) to enrich the training data, we propose to generate data instances inversely by \vtwo{producing} target structures \vtwo{from scratch} first and then prompting the LLM to generate a passage (\ie $Y \rightarrow X$) \vtwo{containing the target structure information}. This inverse design reformulates the synthetic data generation task from structure induction, where models often struggle, to conditional text generation, where LLMs excel. 
\vtwo{
\modelname contains three components.
First, \modelname generates diverse target structures $Y$ from scratch, requiring minimal human efforts to initiate the data generation.
}
In addition, with the ability to customize target structures and control their distribution, we can mitigate data imbalance and improve data diversity by producing target structures encompassing a broader range of triggers, event types and arguments, as well as their various combinations.
Second, \modelname performs instruction-guided data generation to prompt the LLM about the fine-grained task definition and constraints to produce passage $X$. 
Third, \modelname detects errors in the generated data instances via self-reflection and provides hindsight natural language intervention to self-refine the generated data without additional human efforts.

\vtwo{
Experimental results on event extraction (EE) and relation extraction (RE) tasks show that \modelname is capable of generating human-level IE data instances given a couple of exemplar instances as demonstrations without the need for additional ground-truth passage or target structures. For EE, training the supervised model on \modelname-generated data improves the argument classification sub-task by up to 12.91 points in F1 score on the ACE05 dataset, 2.9 points higher than using the same amount of human-curated data. For RE, we observe a 5.41-point F1 score improvement on the TACRED dataset, which is comparable to using the human-curated data. The improvements brought by \modelname-generated data to multiple supervised models across multiple IE tasks demonstrate the generalizability and compatibility of \modelname. Our manual data examination indicates \modelname-produced data exhibits higher passage quality and better aligns with the task definitions compared with human-curated data.

We further conduct a detailed analysis of different methods for using LLMs to improve EE performance, and we show that training the supervised models on \modelname-generated data yields at least 27.51 points higher F1 score for argument classification over the best inference-only LLM formulation.}

Our contributions are three-fold: 1) We introduce \modelname, an inverse structure-to-text IE data generation method with self-refinement; 
\vtwo{
2) We demonstrate the effectiveness, generalizability and compatibility of \modelname by showcasing significant EE and RE performance enhancements and manual data quality examination;
3) We conduct a thorough analysis of different methods for using LLMs in EE, and the most effective approach identified is employing LLMs to generate training data for fine-tuning supervised models.
}

\section{\modelname: Structure-to-Text Data Generation}
\label{sec:method}

We introduce the method design of \modelname as illustrated in \figref{fig:reverse_gen}. \vtwo{We use event extraction (EE) as the exemplar IE task in this section as it covers more method details.} The goal is to create $N$ new data instances ($X$, $Y$) based on $k$ demonstration instances to be used as additional training resources for supervised IE models. Each data instance is composed of a natural language \textbf{passage} $X$ containing event information and a \textbf{target structure} $Y$ containing 0 to any number of \textit{events}, each contains an event trigger, its event type and 0 to any number of (argument mention, argument role) pairs.
There are three steps: 1) target structure generation to prepare $Y$ (\secref{sec:structure}), 2) initial passage generation to generate $X_0$ (\secref{sec:instruction}), and 3) self-refinement with self-reflection to revise $X_0$ to $X_t$ where $t$ is the times of revision (\secref{sec:self-refinement}). 
We introduce the adaptation of \modelname to other IE tasks in \secref{sec:adaptation}.

\subsection{Target Structure Generation}
\label{sec:structure}

The output distribution of the structure prediction dataset largely determines the generalizability and robustness of the model fine-tuned on it. 
We first generate a pool of valid seed words for triggers of each event type, and for arguments of each (event type, argument role) combination. We then create a target structure $Y$. During the process of generating target structures, we particularly employ target structure distribution control to alleviate data imbalance and improve data diversity and comprehensiveness.

\mypar{Trigger candidate generation.} We prompt the LLM with 1) a definition of the selected event type; and 2) a few passages that contain event triggers of the selected event type as demonstrations. We use special tags to wrap the trigger word in the demonstration passages and prompt the LLM to continuously generate more passages with trigger words wrapped within tag pairs. Then we parse the response and extract the trigger candidate words.

\mypar{Argument candidate generation.}
We find reasonable arguments for a certain pair of event type and argument role by prompting LLM with: 1) a definition of the argument role under this specific event type;
2) the entity type we are looking for (\eg a vehicle). The prompt would be like ``\textit{Given the definition of Instrument argument as `The device used to inflict the harm', what are some possible vehicle names that can be used as Instrument?}'' for \textsc{Instrument} argument of \textsc{Life:Injure} event type. The allowed entity types of arguments are provided in an event ontology, and we merge generated word pools returned from separate queries if there are multiple allowed entity types for an argument role. For example, the \textsc{Origin} argument of the \textsc{Movement:Transport} event type could be of entity types \texttt{GPE} (geopolitical entity), \texttt{LOC} (location), or \texttt{FAC} (facility name). \vtwo{We further parse the numbered/bullet lists generated by LLM to get argument candidate words.}

\mypar{Creating target structure and distribution control.}
We randomly sample trigger and argument candidates to create target structure $Y$.
However, unbalanced label distribution and the prevalence of dominant labels pose challenges in many existing human-curated datasets \cite{Ma2023MitigatingBiasQuestion,zhou-etal-2022-sense,cao-etal-2022-intrinsic,zhao-etal-2018-gender}. In EE, a single dominant trigger word eclipses other relevant terms, leading to an unbalanced representation~\cite{Tong2022DocEELargeScaleFinegrained}. We address the issues of imbalanced event type and trigger distribution by evenly generating data instances and significantly expanding the pool size for trigger candidates to 100, which is 1.4$\times$ to 50$\times$ larger for various event types compared to the human-curated ACE05 dataset. Additionally, we balance the \textbf{argument hallucination ratio} in the generated data. This involves ensuring that the generated dataset contains events with both many arguments and few arguments by uniformly replacing argument value with \texttt{None} across different argument hallucination ratios. Furthermore, we balance the \textbf{event density} in passage $X$ by providing target structure $Y$ with 0 to 5 events. Such target structure distribution controls from multiple perspectives ensure the EE models trained on the generated data points effectively learn features and patterns associated with different event and argument densities, enhancing robustness and generalizability.\looseness=-1

\subsection{Instruction-Guided Passage Generation}
\label{sec:instruction}

We use task instruction from multiple task granularities to provide recipes for the LLM to generate passages containing structured event information. The instruction is appended to $k$ in-context learning examples verbalized by our instance verbalizer. Finally, we provide the verbalized target structure information based on the target structure $Y$ to prompt the LLM to generate the initial passage $X_0$.

\mypar{Task-level instruction.} We provide task-related instruction following the annotation guideline curated by experts to guide the human annotation process of the ACE05 dataset~\cite{doddington-etal-2004-automatic}.
Specifically, we provide: 1) a definition of ``event'', ``trigger'', ``participant arguments'' and ``attribute arguments''; 2) an overall task requirement that the goal is to generate a sentence containing the event trigger words and arguments; 3) hallucination clarification that instructs the model \textit{not} to generate arguments of certain roles if we explicitly provide that ``the argument is None''; 4) multiple event clarification that information from multiple events should be contained in a \textit{single} passage.

\mypar{Event type-level instruction.} In this segment, %
we introduce meta-information provided by pre-defined event ontology for a specific event type, including the name and definition of the event type and \textit{each} possible argument roles. \vtwo{We provide all possible argument roles instead of the ones with existing values to ensure the generated passage $X$ does not contain hallucinated arguments that should not appear according to the output structure $Y$. }

\mypar{Instance-level verbalizer.} We verbalize exemplar data instances and target structure $Y$ into natural language sequences with three segments: 1) the number of events in the passage; 2) the content of the event target structure; 3) the passage $X$ with tags wrapping triggers and arguments to explicit hint the LLM about the roles and positions of the keywords, \eg ``\textit{$<$Plaintiff$>$He$<$/Plaintiff$>$ threatened to $<$Trigger$>$sue$<$/Trigger$>$ 
the company.}'' could provide an explicit indicator to the LLM that ``he'' is served as a \textsc{Plaintiff} argument for the event triggered by ``sue''.

\subsection{Self-refinement by Self-reflection}
\label{sec:self-refinement}

After the initial passage $X_0$ is generated, the self-refinement mechanism evaluates the quality and identifies potential errors and further improves the initial generation results through iterative updates~\cite{Madaan2023SelfRefineIterativeRefinementa,Wang2023SelfInstructAligningLanguage,Liu2023ChainHindsightAligns}. 
In the $t$-th refinement iteration, we first identify the potential quality issues of $X_{t-1}$ from a diverse set of quality dimensions (\eg the passage contains \textsc{Crime} argument information, but it should be ``None'' according to $Y$), then the issues are \textbf{feedback} to the LLM by providing a template-based natural language intervention (\eg ``\textit{The passage contains a hallucinated argument \textsc{Crime} incorrectly, remove \textsc{Crime} information for event triggered by `jailed'.}'') along with the generated passage of the previous iteration $X_{t-1}$, so that the LLM could \textbf{refine} the passage and produce $X_t$.

We define a set of quality dimensions and their intervention template manually. 
For EE, they include 
1) whether the trigger/argument mention is a subsequence of the passage;
2) whether a trigger is used to initiate an occurrence; 
3) whether an argument is used as an event participant or attribute of the specific event; 
4) whether the argument is serving the required argument role; 
5) whether the passage contains information that could serve as an argument that should not appear; 
6) whether POS tags of the argument mentions in the passage context match the provided ones.

For each quality dimension, we query LLM with questions like ``\textit{Is  `Syria' a \textsc{Destination} argument describing the event triggered by `flee'?}''. We then standardize LLM's response to a binary error identification flag by checking whether the response entails a confirmative phrase ``Yes, it is.'' with a Natural Language Inference model fine-tuned on MultiNLI~\cite{Williams2018BroadCoverageChallengeCorpus} based on BART-large~\cite{Lewis2020BARTDenoisingSequencetoSequencea}. If a quality issue is flagged, we use the intervention template corresponding to the selected quality dimension as part of the feedback to the LLM for iterative revision. Such a self-reflection design makes the self-refinement process generalizable and robust since the entire error identification process through self-reflection and the revision process are done by the LLM itself without external add-on components.\looseness=-1

\vtwo{
\subsection{Adaptation to Relation Extraction}
\label{sec:adaptation}

RE's relation type would be the equivalent concept of ``event type'' in EE.
For target structure generation, we generate entity candidates using seed data instances' entities as in-context examples. We then randomly pair entity candidates and assign a relation between the two entities. For initial passage generation, we use relation type definition instead. For self-refinement, we use the quality dimensions: 1) whether the given entities are contained in the generated passage, 2) whether there is a relation between them, and 3) whether they hold the certain relation provided in $Y$. 
}

\section{Experiments on Event Extraction}
\label{sec:eval}
To evaluate the efficiency of \modelname-generated data for event extraction, we compare the performance of supervised EE models with and without using the \modelname-generated data. 

\subsection{Baselines}
We use two types of models as our baselines: the \textit{inference-only methods}, and the \textit{supervised models} fine-tuned on data \vtwo{created} by various \textit{\vtwo{data creation strategies}}.\looseness=-1

\mypar{Inference-only EE methods.} 
We use LLM GPT-3.5~\cite{chatgpt} and GPT-4~\cite{gpt4} to perform inference.\footnote{\vtwo{We use \texttt{gpt-3.5-turbo-0301} and \texttt{gpt-4-0314}.}} We adopt different EE input-target formulations to prompt LLMs, including formulations inspired by generative supervised models (1-3) and LLM prompting methods specifically designed for EE proposed by recent works (4-6). The formulations include:
1) \textbf{Examples \& IO (Text2Event)}~\cite{Lu2021Text2EventControllableSequencetoStructure} uses a concise but \textit{unnatural} template to represent event structure.
2) \textbf{Examples \& IO (DEGREE)}~\cite{Hsu2022DEGREEDataEfficientGenerationBased} generates a filled-in \textit{natural} language template. 3) \textbf{Examples \& IO (DICE)}~\cite{Ma2023DICEDataEfficientClinical} is similar to DEGREE but uses separate queries for different argument roles. 4) \textbf{Task Instruction}~\cite{Li2023EvaluatingChatGPTInformation} provides task description and pre-defined event type \textit{names}. 5) \textbf{Instruction+Examples}~\cite{Gao2023ExploringFeasibilityChatGPT} provides event type \textit{definitions} and positive and negative examples, in addition to the task description. 6) \textbf{Code4Struct}~\cite{Wang2023Code4StructCodeGeneration} formulates task definition, event type definition and examples in Python code. 
For baselines 1-3, we follow the original input and target formulations and additionally provide $k$ demonstration input-target pairs contained in the input prompt for in-context learning. Baselines 4 and 5 only support Tri-I and Tri-C, and baseline 6 only supports Arg-I and Arg-C.\looseness=-1

\mypar{Supervised EE models.} We use two representative EE models as the testbed to evaluate the quality of the generated data. 
\textbf{OneIE}~\cite{Lin2020JointNeuralModel} is a multi-task sequence-tagging model trained with global features based on RoBERTa-large~\cite{Liu2019RoBERTaRobustlyOptimized}. Note that we remove the human-labeled data for tasks other than EE (\eg entity identification, relation extraction) to enable a fair comparison. \textbf{DEGREE}~\cite{Hsu2022DEGREEDataEfficientGenerationBased} is a prompt-based model that fills in event type-specific human written templates based on a BART-large pre-trained model~\cite{Lewis2020BARTDenoisingSequencetoSequencea}.\looseness=-1

\begin{table*}[th]
\begin{center}
{
\small
\setlength\tabcolsep{4.7pt}
\begin{tabular}{rll|rHrr|rHrr|rHrr|rHrr}
\toprule
\multirow{2}{*}[-4pt]{\#} &
\multirow{2}{*}[-4pt]{\makecell[l]{}} &
\multirow{2}{*}[-4pt]{\makecell[l]{}} & $k=0$ & $1$ & $5$ & $10$ & $k=0$ & $1$ & $5$ & $10$ & $k=0$ & $1$ & $5$ & $10$ & $k=0$ & $1$ & $5$ & $10$
\\ \cmidrule{4-19}
& &  & \multicolumn{4}{c|}{Trigger Iden.} & \multicolumn{4}{c|}{Trigger Clas.} & \multicolumn{4}{c|}{Argument Iden.} & \multicolumn{4}{c}{Argument Clas.}
\\ \toprule
\multicolumn{19}{c}{\cellcolor{blue!10}\textbf{\textit{Inference-only Methods}}}
\\
& \textit{LLM} & \textit{Formulation}
\\ \midrule
1 & \multirow{6}{*}{GPT-3.5}  & E\&IO (Text2Event)
& 0.00 & & 9.23 & 11.30 & 0.00 & & 2.12 & 3.47
& 0.00 & & 0.87 & 1.03 & 0.00 & & 0.31 & 0.44
\\
2 & & E\&IO (DEGREE)
& 0.00 & & 14.39 & 17.52 & 0.00 & & 3.17 & 6.21
& 0.00 & & 1.02 & 2.47 & 0.00 & & 0.92 & 1.98
\\
3 & & E\&IO (DICE)
& 0.00 & & 15.13 & 16.94 & 0.00 & & 4.11 & 7.09
& 0.00 & & 0.71 & 1.65 & 0.00 & & 0.33 & 0.97
\\
4 & & Task Inst.$^{\mathsection}$
& 18.31 & 18.31 & 18.31 & 18.31 & 8.37 & 8.37 & 8.37 & 8.37
& \multicolumn{4}{c|}{---} & \multicolumn{4}{c}{---}
\\
5 & & Inst.+Examples
& 29.44 & 00.00 & 47.24 & 59.71 & 21.56 & 00.00 & 40.57 & 53.29
& \multicolumn{4}{c|}{---} & \multicolumn{4}{c}{---}
\\
6 & & Code4Struct
& \multicolumn{4}{c|}{---} & \multicolumn{4}{c}{---} 
& 12.33 & & 18.34 & 23.74 & 9.72 & & 14.85 & 19.10
\\ \midrule
7 & \multirow{2}{*}{GPT-4} & Inst.+Examples
& \textbf{34.31} & 00.00 & \textbf{52.55} & \textbf{62.12}
& \textbf{27.35} & 00.00 & \textbf{46.57} & \textbf{56.46}
& \multicolumn{4}{c|}{---} & \multicolumn{4}{c}{---}
\\
8 & & Code4Struct
& \multicolumn{4}{c|}{---} & \multicolumn{4}{c|}{---} 
& \textbf{17.51} & & \textbf{24.50} & \textbf{27.62}
& \textbf{11.89} & & \textbf{24.28} & \textbf{25.48}
\\
\toprule
\multicolumn{19}{c}{\cellcolor{blue!10}\textbf{\textit{Supervised Models}} ($N=50$ except line 9 \& 14)}
\\ 
& \textit{EE Model} & \textit{Data Creation}
\\ \midrule
9 & \multirow{5}{*}{OneIE} & None ($N=0$)
& \vtwo{0.00} &  & \vtwo{57.24} & 60.55
& \vtwo{0.00} &  & \vtwo{52.38} & 54.84 
& \vtwo{0.00} &  & \vtwo{29.06} & 36.45 
& \vtwo{0.00} &  & \vtwo{25.85} & 33.56 
\\
10 & & Weak Sup.
& 29.48 & & 49.23 & 51.66
& 23.61 & & 45.02 & 45.23
& 16.19 & & 24.35 & 26.84
& 10.47 & & 19.14 & 22.94
\\
11 & & \modelname (GPT-3.5)
& 42.61 && 63.08 & 64.12 & 36.65 && 56.61 & 57.29
& 30.32 && 39.76 & 43.40 & 24.36 && 36.17 & 40.93
\\
12 & & \modelname (GPT-4)
& \textbf{45.42} && \textbf{64.63} & \cellcolor{green!10}\textbf{\underline{66.77}} & \textbf{39.15} && \textbf{58.84} & \cellcolor{green!10}\textbf{\underline{60.76}}
& \textbf{32.23} && \textbf{42.76} & \cellcolor{green!10}\textbf{\underline{46.22}} & \textbf{27.47} && \textbf{39.53} & \cellcolor{green!10}\textbf{\underline{43.25}}
\\
13 & & \cellcolor{gray!15}Human$^{\dag\mathsection}$
& \cellcolor{gray!15}65.62 & \cellcolor{gray!15}65.62 & \cellcolor{gray!15}65.62 & \cellcolor{gray!15}65.62 
& \cellcolor{gray!15}60.10 & \cellcolor{gray!15}60.10 & \cellcolor{gray!15}60.10 & \cellcolor{gray!15}60.10 
& \cellcolor{gray!15}44.76 & \cellcolor{gray!15}44.76 & \cellcolor{gray!15}44.76 & \cellcolor{gray!15}44.76 
& \cellcolor{gray!15}41.60 & \cellcolor{gray!15}41.60 & \cellcolor{gray!15}41.60 & \cellcolor{gray!15}41.60 
\\ 
\midrule
14 & \multirow{5}{*}{DEGREE} & None ($N=0$)
& \vtwo{0.00} &  & \vtwo{55.62} & 57.65 
& \vtwo{0.00} &  & \vtwo{50.69} & 52.49
& \vtwo{0.00} &  & \vtwo{31.77} & 42.29
& \vtwo{0.00} &  & \vtwo{30.19} & 40.08 
\\
15 & & Weak Sup.
& 27.51 & & 46.48 & 49.70
& 22.23 & & 41.65 & 43.55
& 18.14 & & 32.53 & 33.33
& 13.45 & & 27.38 & 30.01
\\
16 & & \modelname (GPT-3.5)
& 43.74 && 61.39 & \cellcolor{green!10}\underline{63.57} & 38.90 && 56.41 & \cellcolor{green!10}\underline{59.10}
& 32.32 && 48.73 & \cellcolor{green!10}\underline{53.06} & 28.21 && 46.55 & \cellcolor{green!10}\underline{50.97}
\\
17 & & \modelname (GPT-4)
& \textbf{46.69} & & \cellcolor{green!10}\textbf{\underline{64.47}} & \cellcolor{green!10}\textbf{\underline{65.17}}
& \textbf{41.75} & & \cellcolor{green!10}\textbf{\underline{59.92}} & \cellcolor{green!10}\textbf{\underline{61.42}}
& \textbf{35.85} & & \textbf{51.92} & \cellcolor{green!10}\textbf{\underline{54.56}}
& \textbf{32.09} & & \cellcolor{green!10}\textbf{\underline{50.74}} & \cellcolor{green!10}\textbf{\underline{52.99}}
\\
18 & & \cellcolor{gray!15}Human$^{\dag\mathsection}$
& \cellcolor{gray!15}63.49 & \cellcolor{gray!15}63.49 & \cellcolor{gray!15}63.49 & \cellcolor{gray!15}63.49 
& \cellcolor{gray!15}58.86 & \cellcolor{gray!15}58.86 & \cellcolor{gray!15}58.86 & \cellcolor{gray!15}58.86 
& \cellcolor{gray!15}52.47 & \cellcolor{gray!15}52.47 & \cellcolor{gray!15}52.47 & \cellcolor{gray!15}52.47 
& \cellcolor{gray!15}50.09 & \cellcolor{gray!15}50.09 & \cellcolor{gray!15}50.09 & \cellcolor{gray!15}50.09 
\\
\bottomrule
\end{tabular}
}
\end{center}
\caption{ 
Event extraction performance (F1, \%) when using $k$ seed data instances per event type.
\vtwo{
\textit{Inference-only methods} and \textit{data creation methods} use the same set of $k$ examples for each event type to prompt LLM to perform EE and generate data instances respectively. \textit{Supervised EE models} are trained on $k + N$ data instances per event type.
}
\textbf{Boldface} indicates the best performance among each group (line 1-8, 9-13 and 14-18) without additional human efforts. 
$\dag$ and \colorbox{gray!20}{gray} background indicate 
\vtwo{using ACE05 human-curated data sampled, thus it is not comparable with other lines.}
\underline{Underlining} and \colorbox{green!10}{green} background indicate \modelname-generated data improve EE performance more than human-curated data. 
There is no difference when using various $k$ for lines indicated by $\mathsection$ as no seed data instances are used. 
We use the trigger and event type produced by the best upstream model (\ie Inst.+Examples) as inputs to Code4Struct to compensate for its lack of event detection capabilities.
}
\label{table:overall_k}
\end{table*}

\mypar{\vtwo{Data creation strategies.}} 
Besides \modelname, we introduce two other approaches to obtain training data. \textbf{Weakly Supervision}: we use the best inference-only model for EE (\ie Inst.+Examples for Tri-I and Tri-C, and Code4Struct for Arg-I and Arg-C) to predict event structure $Y^\prime$ from passage $X$, and the ($X$, $Y^\prime$) pairs are used as training data. \textbf{Human}: we use human-curated data instances randomly sampled from the ACE05 dataset\vtwo{, which requires much more human annotators' efforts, as an ideal but unrealistic setting}.

\subsection{Experimental Setup}
We denote $k$ as the number of demonstration examples of each event type used as in-context demonstrations for the inference-only methods and \vtwo{data creation strategies}, and $N$ as the number of data instances per event type \vtwo{created by data creation strategies}.
Supervised models are trained on $k+N$ data instances per event type. 
We use the full test set for evaluation. We use the event ontology and data instances from the widely used sentence-level English event extraction dataset ACE05 \cite{doddington-etal-2004-automatic}. 

We follow previous EE works~\cite{Lin2020JointNeuralModel} and report 
F1 scores for four tasks.\footnote{The extracted trigger/argument has to match the ground truth \textit{exactly}, instead of head word matching or coreference matching used in works~ \cite{Li2021DocumentLevelEventArgument,Wang2023Code4StructCodeGeneration}.}
1) Trigger Identification: identified trigger span is correct.
2) Trigger Classification: its predicted \textit{event type} is also correct.
3) Argument Identification: identified argument span is correct.
4) Argument Classification: its predicted \textit{argument role} is also correct. Note that each task is dependent on the output of the previous task. \vtwo{We report the medium result for three runs of different random seeds.}

\subsection{\vtwo{Effectiveness of Data Generation}}

\begin{figure*}[h!]
\centering
  \includegraphics[width=\textwidth]{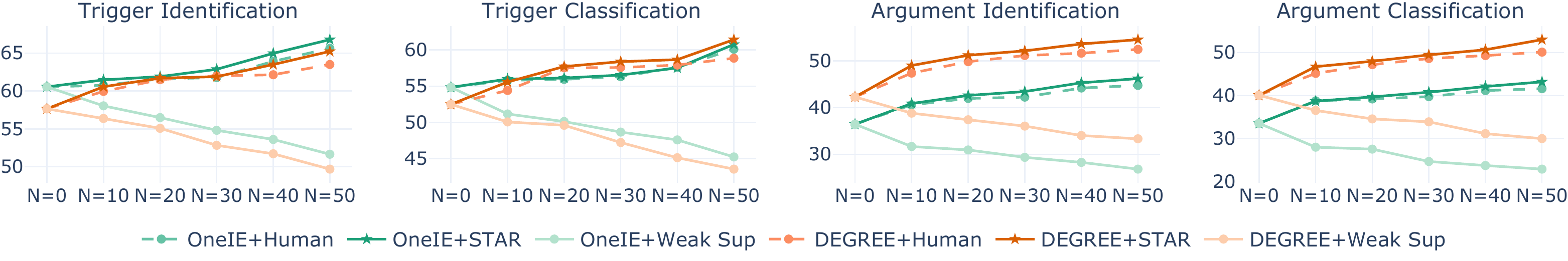}
\caption{
Event extraction performance (F1, \%) when the EE models are trained on $N$ augmented training data on top of 10 data points ($k=10$) for each event type. 
\vtwo{
We observe that performance gain brought by \modelname-generated data is magnified as the data augmentation scales up with a larger $N$,}
and data generated by \modelname is even more effective than human-curated ones. We use GPT-3.5 version \modelname for this set of experiments. 
}
\label{fig:overall_N}
\end{figure*}

\tbref{table:overall_k} shows the EE performance when using various numbers ($k$) of demonstrations and \figref{fig:overall_N} further shows the effects of various amounts ($N$) of augmented data instances.

\mypar{\modelname boosts low-resource EE performance.} \tbref{table:overall_k} shows that data generated by \modelname significantly improve the supervised models' performance (line 9 vs \vtwo{11-12}, and line 14 vs 16-17) across all tasks. The F1 scores of Arg-C are improved by 9.69 and 12.91 respectively for OneIE and DEGREE\vtwo{ when $k=10$}. 

\mypar{Data produced by \modelname is more effective than human-curated ones given sufficient examples.} Compared to human-curated data instances (line 13 and 18 in \tbref{table:overall_k}), training supervised models with \modelname-generated data leads to better performance when 5 or more demonstrations are used for \modelname
and the supervised model is DEGREE except for Arg-I (indicated by \underline{underlined} results in line 17). We also observe a similar trend for OneIE when 10 demonstrations are used for \modelname (\underline{underlined} results in line 12).
This indicates we could boost low-resource EE performance as if we had additional ground-truth data without paying the human annotation efforts. \figref{fig:overall_N} further shows the superiority of \modelname-generated data over human-curated ones regardless of the number of augmented data instances ($N$) using 10 demonstrations ($k=10$) for all supervised models.\looseness=-1

\mypar{Noisy data generated by weak supervision impairs performance.} 
The performance of the best existing few-shot method \vtwo{used for} weakly supervised data generation is still not satisfactory (line 7-8), leading to poor data quality and further impairing the performance of the supervised models by at least 10 points decline in F1 scores for Arg-C (line 9/14 vs 10/15). Shown by the downward curves in \figref{fig:overall_N}, the performance decay is larger when more noisy data is used. This demonstrates that data generation methods that induce semi-supervised label $Y^\prime$ from $X$ tend to perform poorly for complicated IE tasks like EE.

\subsection{\vtwo{Best Recipe to Use LLM for EE}}

\mypar{Tunable supervised models are much better than inference-only methods.} Supervised models (without data augmentation) yield a 14.6-point higher F1 score than the best inference-only method (line 8 vs 14) for Arg-C \vtwo{when $k=10$}. 
The inference-only methods do not fully utilize the limited resources because they cannot learn from given demonstrations across all types. They suffer from the limitations of understanding the target structure from task description or inducing such patterns from in-context examples. We observe that supervised methods are better in handling low-resource EE, and their performances are further improved by training on additional data generated by \modelname.

\mypar{When prompting LLM for EE, examples are crucial and code formulation helps.} Among various formulations for the inference-only methods, we observe that the natural language template (used in DEGREE) is more beneficial than the unnatural one (used in Text2Event) due to its better utilization of the pertaining knowledge. We also observe the importance of considering dependencies among arguments.
This is evidenced by the superior performance of DEGREE over DICE, which queries each argument role separately for argument extraction. The low scores of Text2Event, DEGREE, and DICE, which all use examples \textit{without} task description, highlight the importance of describing requirements and rules in the instruction. It is worth noting that there is a substantial 45-point increase in F1 score from ``Task Inst.'' to ``Inst.+Examples'' when $k=10$ in the trigger classification task. 
This result highlights the effectiveness of the in-context examples in providing the LLM with the target structure knowledge. Additionally, we note that code formulation outperforms natural language, aligning with findings from previous studies~\cite{Wang2023Code4StructCodeGeneration}.

\subsection{Effects of Demonstration Quantity $k$ and Data Quantity $N$}

We observe that demonstration examples for the data generation pipeline are crucially important to bring in target structure information and showcase the task constraints in action. \vtwo{When no demonstration example is provided (\ie $k=0$), the data generated by \modelname could still significantly boost the supervised models' performance, but it is much worse than training with human-curated data (line 12/17 vs 13/18 of \tbref{table:overall_k}).}
This indicates that even the state-of-the-art LLM GPT-4 is not capable of generating human-level EE data instances without demonstrations.
The more demonstrations used in the data generation pipeline, the better the performance. \figref{fig:overall_N} shows that the performance gain brought by augmented data is magnified as the data augmentation scales up. Due to resource limitation, we use a maximum $k$ of 10 and $N$ of 50 in the experiments, and we anticipate the EE performance to further improve with larger $k$ and/or $N$.

\begin{table}[th]
\begin{center}
{
\small
\setlength\tabcolsep{5pt}
\begin{tabular}{cl|cccc}
\toprule
\# & Method Variant & Tri-I & Tri-C & Arg-I & Arg-C
\\ \midrule
\multicolumn{6}{c}{Target Structure $Y$ Generation Methods} \\
\midrule
1 & \cellcolor{gray!15}Ground-truth $Y$\textsuperscript{\dag}
& \cellcolor{gray!15}59.21 & \cellcolor{gray!15}54.03 & \cellcolor{gray!15}44.13 & \cellcolor{gray!15}41.77
\\
2 & LLM generation
& 60.52 & 54.80 & 48.00 & 45.73
\\ \midrule
\multicolumn{6}{c}{Error Identification Strategies} \\
\midrule
3 & None
& 58.33 & 52.94 & 42.18 & 39.85
\\
4 & Rule-based checking
& 59.42 & 53.37 & 43.77 & 41.26
\\ 
5 & Self-reflection (NLI)
& 59.89 & 54.02 & 46.11 & 43.54
\\
6 & Self-reflection (LLM)
& 60.52 & 54.80 & 48.00 & 45.73
\\
\bottomrule
\end{tabular}
}
\caption{
Ablation study on DEGREE's EE results while using 10 demonstrations with 10 additional generated data instances per event type ($k=10$, $N=10$).
}
\label{table:ablation}
\end{center}
\end{table}

\subsection{Ablation Studies}
We investigate the influence of the design choices in \tbref{table:ablation}. 
\mypar{Structure generation method.} The diversity and balanced distribution of the generated target structures produced by our target structure generation component (\secref{sec:structure}) result in an almost 4-point higher \vtwo{Arg-C} F1 score compared to using human-annotated target structures sampled from the ACE05 dataset. Our target structure generation component yields highly diverse structures $Y$ without guaranteeing their factuality and commonsensical nature, suggesting that diversity outweighs factuality in terms of impact on performance. 

\mypar{Error identification strategy.} We also investigate the error identification capabilities of our self-reflection module with LLM as the backbone (\secref{sec:self-refinement}). \vtwo{We compare it with two alternative methods and we utilize the same template to provide feedback on the identified errors.
\textbf{Rule-based checking} uses heuristics to check whether a trigger/argument is a subsequence of the generated passage and uses an external NER module~\cite{Yamada2020LUKEDeepContextualized} to check whether a trigger/argument functions as the desired entity type. \textbf{Self-reflection (NLI)} uses the generated passage as the premier and a statement of a quality dimension as the hypothesis. We use entailment prediction of the NLI module used in \secref{sec:self-refinement} to identify whether a certain quality issue exists.
The results are in lines 4-6 of \tbref{table:ablation}. 
}
Our observations demonstrate both alternative methods help (line 3 vs 4-5), and self-reflection with LLM exhibits the highest effectiveness in error identification (line 6). Notably, the results underscore the effectiveness of the self-reflection design, resulting in a substantial 6-point increase in the F1 score for Arg-C without the need for additional annotation efforts.\looseness=-1

\section{Experiments on Relation Extraction}

\begin{table}[th]
\begin{center}
{
\small
\setlength\tabcolsep{5pt}
\begin{tabular}{cll|ccc}
\toprule
\# & RE Model & Data Gen & $N=0$ & $10$ & $40$ 
\\ \midrule
1 & GPT-3.5 & ---
& 27.91 & 27.91 & 27.91
\\ \midrule
2 & \multirow{3}{*}{SURE} & Weak Sup.
& \multirow{3}{*}{27.61} &28.02&28.32
\\
3 &  & \modelname (GPT-3.5)&&\cellcolor{green!10}\textbf{\underline{30.50}} & \textbf{33.02}
\\
4 &  & \cellcolor{gray!15}Human\textsuperscript{\dag} &&\cellcolor{gray!15}30.11 & \cellcolor{gray!15}35.62 
\\ \midrule 
5 & \multirow{3}{*}{GenPT} & Weak Sup. 
& \multirow{3}{*}{33.38} & 30.93 & 30.29
\\
6 & & \modelname (GPT-3.5) &&\textbf{34.55}&\textbf{37.01}
\\
7 & & \cellcolor{gray!15}Human\textsuperscript{\dag} & & \cellcolor{gray!15}36.74& \cellcolor{gray!15}37.61
\\
\bottomrule
\end{tabular}
}
\end{center}
\caption{
Relation extraction performance (\%) when the RE models are trained on $N$ augmented training data on top of 10 seed data instances ($k=10$) for each relation type.
}
\label{tb:overall_RE}
\end{table}

To assess the generalizability of our proposed method, we conduct experiments on the sentence-level relation extraction (RE) task. \vtwo{We use relation definitions and seed examples in the widely-used TACRED dataset~\cite{Zhang2017PositionawareAttentionSupervised}.} In this task, we generate (subject, relation, object) tuples from scratch,  providing additional training data. The RE task aims to identify the relation between the given subject and object entities within a context passage.
We train two representative RE models on the generated data instances. \textbf{SURE}~\cite{Lu2022SummarizationIndirectSupervision} converts the task into a summarization formulation to leverage the indirect supervision with PEGASUS-large as the pre-trained encoder~\cite{zhang2020pegasus}. \textbf{GenPT}~\cite{han2022generative} transforms RE into an infilling problem with a RoBERTa-large model as backbone~\cite{Liu2019RoBERTaRobustlyOptimized}.
We use the same set of data \vtwo{creation} baselines as in the EE experiments. 
\vtwo{
For the weakly supervised baseline, we use SURE's formulation by querying the LLM once for each relation label, and the model outputs whether the verbalized relation between head and tail entities is entailed by the passage with $k$ additional in-context examples, which is the most empirically superior RE formulation in our exploration.
}
We report micro F1 score across all relations (except for the ``no relation'' class) following prior works~\cite{Lu2022SummarizationIndirectSupervision,lu-etal-2023-multi}.\looseness=-1

\tbref{tb:overall_RE} presents the RE performance.
The \modelname-generated data significantly enhances the performance compared to $N=0$, with improvements of 5.4 and 3.6 \vtwo{F1 points} when using SURE and GenPT respectively.

\section{Quality Verification}
\vtwo{
Two annotators who are familiar with the EE task manually assess the quality of the EE data generated by \modelname and curated by humans sampled from the ACE05 dataset in terms of 3 passage-related quality dimensions (grammatical, informative and commonsensical levels), and 5 task-specific quality dimensions (whether the trigger, event type, argument, and argument role provided in the generated target structure $Y$ is being used in the generated passage $X$ following their definitions) for data instances in 100 sentences. We report the average scores of the two annotators. 

\tbref{table:human_anno} shows both sets of data demonstrate high satisfactory levels. \modelname-generated data exhibits higher passage quality and better follows the task definition for most metrics, suggesting \modelname produces EE annotations with comparable or even better quality than human annotators.\looseness=-1
}
\begin{table}[th]
\begin{center}
{
\small
\setlength\tabcolsep{3.8pt}
\begin{tabular}{Hl|rHr}
\toprule
\# & Quality Dimension & \modelname & Inst. & Human
\\ \midrule
& Grammaticality of the generated passage ($X$)
& \textbf{96} & & 90
\\
& Informativeness of the generated passage ($X$)
& \textbf{79} & & 78
\\
& Commonsense of the generated passage ($X$)
& \textbf{95} & & 93
\\ \midrule
& Trigger span describes event occurrence
& \textbf{99} & & \textbf{99}
\\
& Event follows event type definition
& \textbf{99} & & 97
\\
& Argument span describes an event
& \textbf{100} & & 99
\\
& Argument associated with correct trigger
& \textbf{98} & & 95
\\
& Argument follows role definition
& 98 & & \textbf{99}
\\
\bottomrule
\end{tabular}
}
\caption{
\vtwo{Human quality assessment satisfactory rate (\%).\looseness=-1}
}
\label{table:human_anno}
\end{center}
\end{table}

\section{Related Works}
\label{sec:relatedworks}

\subsection{Low-Resource \vtwo{Information} Extraction}

Existing low-resource IE methods mostly borrow supervision signals from other tasks, such as Abstract Meaning Representation~\cite{Huang2018ZeroShotTransferLearning} or Semantic Role Labeling~\cite{Zhang2021ZeroshotLabelAwareEvent}, via transfer learning or reformulate the task as other data-rich tasks, such as Natural Language Inference (NLI)~\cite{Xu2023CanNLIProvidea,Sainz2022TextualEntailmentEvent}, summarization~\cite{Lu2022SummarizationIndirectSupervision} or QA~\cite{Lyu2021ZeroshotEventExtraction,ma-etal-2023-parameter}, via indirect supervision. 
Recent works explore prompting LLM with task instruction and examples to predict event structure from input passage by prompting task requirement~\cite{Li2023EvaluatingChatGPTInformation}, or providing examples along with task instruction~\cite{Gao2023ExploringFeasibilityChatGPT,Xu2023InstructionsBackdoorsBackdoor}.
However, their performance is much worse than fine-tuning a model with training data\vtwo{ for complicated IE tasks like EE}, which motivates us to focus on \vtwo{generating training data for supervised IE models.}\looseness=-1

\subsection{Data Generation}

Existing works perform data augmentation by producing analogous input ($X_{gold} \rightarrow X^\prime$)~\cite{Kumar2020DataAugmentationUsing,lee2021neural}, and annotating unlabeled examples with existing models ($X_{gold} \rightarrow Y$)~\cite{He2021GenerateAnnotateLearn} on tasks like textual similarity~\cite{Schick2021GeneratingDatasetsPretrained}, relation extraction~\cite{Chia2022RelationPromptLeveragingPromptsa}, text classification, QA, NLI~\cite{Ye2022ZeroGenEfficientZeroshot} and more~\cite{Wang2023SelfInstructAligningLanguage,Tang2023DoesSyntheticData}.
Another line of prior work assumes that the labels are given and it prompts an LLM to generate input ($Y_{gold} \rightarrow X$), such as given sentiment labels~\cite{Meng2022GeneratingTrainingData,Gao2021MakingPretrainedLanguage} or relation triples~\cite{Josifoski2023ExploitingAsymmetrySynthetic}. 
However, these works generate data for tasks with significantly simpler output structures than \vtwo{many IE tasks like EE}. Furthermore, they require existing ground-truth label data to perform data generation while our work aims to generate both $Y$ and $X$ from scratch.

\section{Conclusion}
\label{sec:conclusion}
We present \modelname, an inverse data generation pipeline designed for low-resource IE that generates output structure first and then curates input passage containing structure content
with self-refinement capabilities to fix self-identified error cases.
Experimental results on EE and RE show the generated data instances significantly improve the performance.\looseness=-1

\section*{Acknowledgments}

Utkarsh Lal and Michael M. Song made significant contributions to the quality verification section. Many thanks to lab members at UCLA SCAI and UCLANLP for their suggestions, and to the anonymous reviewers for their feedback. 
This effort was sponsored by the Defense Advanced Research Project Agency (DARPA) grant HR00112290103 and AFOSR MURI via grant \#FA9550-22-1-0380.

\bibliography{ma_auto,custom,anthology_small}

\end{document}